\documentclass[sigplan,screen]{acmart}
\usepackage{multirow}
\usepackage{url}
\usepackage{algorithmic}
\usepackage{algorithm2e}

\usepackage{subfigure}
\usepackage{pifont}
\usepackage[section]{placeins}

\AtBeginDocument{%
  \providecommand\BibTeX{{%
    \normalfont B\kern-0.5em{\scshape i\kern-0.25em b}\kern-0.8em\TeX}}}

\setcopyright{acmcopyright}
\copyrightyear{2023}
\acmYear{2023}
\acmDOI{XXXXXXX.XXXXXXX}

\acmConference[MM '23]{Proceedings of the 31st ACM International Conference on Multimedia}{October 29 – November 3, 2023}{Ottawa, Canada}
\acmPrice{15.00}
\acmISBN{978-1-4503-XXXX-X/18/06}

\acmSubmissionID{326}

\begin{document}

\title{AmGCL: Feature Imputation of Attribute Missing Graph via Self-supervised Contrastive Learning}

\author{Xiaochuan Zhang}
\authornote{Both authors contributed equally to this research.}
\email{zhangxiaochuan-jk@360shuke.com}
\author{Mengran Li}
\authornotemark[1]
\email{limengran-jk@360shuke.com}
\affiliation{%
  \institution{Qifu Technology, Inc.}
  \state{Beijing}
  \country{China}
}


\author{Ye Wang}
\email{wangye3-jk@360shuke.com}
\affiliation{%
  \institution{Qifu Technology, Inc.}
  \state{Beijing}
  \country{China}
}

\author{Haojun Fei}
\email{feihaojun-jk@360shuke.com}
\affiliation{%
  \institution{Qifu Technology, Inc.}
  \state{Beijing}
  \country{China}
}






\begin{abstract}
Attribute graphs are ubiquitous in multimedia applications, and graph representation learning (GRL) has been successful in analyzing attribute graph data. However, incomplete graph data and missing node attributes can have a negative impact on media knowledge discovery. Existing methods for handling attribute missing graph have limited assumptions or fail to capture complex attribute-graph dependencies. To address these challenges, we propose Attribute missing Graph Contrastive Learning (AmGCL), a framework for handling missing node attributes in attribute graph data. AmGCL leverages Dirichlet energy minimization-based feature precoding to encode in missing attributes and a self-supervised Graph Augmentation Contrastive Learning Structure (GACLS) to learn latent variables from the encoded-in data. Specifically, AmGCL utilizies feature reconstruction based on structure-attribute energy minimization while maximizes the lower bound of evidence for latent representation mutual information. Our experimental results on multiple real-world datasets demonstrate that AmGCL outperforms state-of-the-art methods in both feature imputation and node classification tasks, indicating the effectiveness of our proposed method in real-world attribute graph analysis tasks. 
\end{abstract}

\ccsdesc[500]{Computing methodologies~Knowledge representation and reasoning}
\ccsdesc[500]{Computing methodologies~Neural networks}

\keywords{Attribute missing graph, graph neural network, missing data imputation, self-supervised contrastive learning, node classification}


\received{20 February 2007}
\received[revised]{12 March 2009}
\received[accepted]{5 June 2009}

\maketitle

\section{Introduction}

Graph structures are a powerful way to model relationships between entities in many real-world media applications, such as social networks (e.g., Facebook, Twitter) \cite{myers2014information}, financial businesses (e.g., stock market analysis) \cite{cheng2022financial}, and recommendation systems (e.g., Netflix, Spotify) \cite{he2020lightgcn}. The edges of the graph represent complex interaction relationships, while nodes carry rich information representation. With the advent of deep learning, numerous graph neural network (GNN) models have been developed to enable graph representation learning. The goal of graph representation learning is to learn latent vector representations of nodes that can be used for downstream tasks such as node classification \cite{kipf2016semi} and link prediction \cite{zhang2018link}. These tasks are important for various applications in industry and academia. For example, in media network analysis, node classification can help identify influential users or detect communities, while link prediction can help recommend friends or suggest new connections.

In real-life scenarios, graphs often include node attributes, which are known as attribute graphs and provide additional information about the nodes \cite{you2020handling, liu2020attribute}. However, real-world attribute graphs often suffer from missing attributes, as some users may set their profiles to private in social networks, and users in financial lending may register without providing descriptive information \cite{krasnova2010online}. This creates a challenge for attribute-missing graph representation learning, which needs to effectively utilize all available node features despite missing and unbalanced semantic information. 

Conventional methods for handling missing node attributes, such as mean and median imputation \cite{xia2017adjusted, you2020handling}, and k-nearest neighbor imputation \cite{silva2015single}, rely on using statistical measures from observed data to replace missing values. However, these methods may introduce bias and adversely affect downstream tasks in the face of non-uniformly missing data. Moreover, these methods only consider local information of nodes and do not take into account the global information of nodes in the graph. In contrast, graph neural network-based methods such as GCN \cite{kipf2016semi} and GAT \cite{velivckovic2017graph} allow for the direct inclusion of missing values during training and take into account the global information of nodes in the graph. However, these methods may not be specifically designed for attribute-missing graphs and may fail to accurately interpolate missing values or learn informative representations for downstream tasks, as they do not explicitly consider the relationships between missing attributes and other attributes in the graph structure. Therefore, developing methods that can effectively utilize both local and global information while addressing the issue of missing attribute data remains an open research challenge.

Recent research has explored using generative GNN-based methods to complete attribute missing graph representation learning \cite{garcia2019combining}. These methods assume latent variables for missing attribute features, fit the real distribution, and fill in the features using techniques such as GAN \cite{chen2022learning} and GMRF \cite{yoo2022accurate}.  They then use the relationship between attribute information and structural information to learn the latent representation and complete downstream tasks. However, they lack a mechanism for supervised learning of latent representation, which may lead to poor performance in downstream tasks. Additionally, the performance of the model largely depends on the quality of the learned latent representations, and the quality of the generated representations may not always be reliable. Therefore, in the absence of effective constraints on the latent representations, these generative methods may fail to capture the complex relationships between attributes and the underlying graph structure, resulting in suboptimal performance.

In this paper, we propose a novel method, AmGCL, for learning graph representations with attribute missing using a graph contrastive learning framework. AmGCL aims to create an informative latent representation that is minimally influenced by the missing input features. To achieve this, we use a Dirichlet energy minimization-based precoder \cite{jeffreys1999methods} to precode missing attribute features. Then, we feed both attribute and structural features into the Graph Augmentation Contrastive Learning Structure (GACLS), which employs two distinct encoders to generate latent representations from the augmented attribute missing graphs. The target generator is responsible for feature induction. The encoder networks in GACLS are trained to match the output of the target generator at the missing nodes. The target generator acts as a teacher network in a self-knowledge refinement or contrastive methods \cite{shi2021semi}. AmGCL is designed for feature induction and downstream tasks and does not rely on complex constraints for generating potential representations.
In summary, this paper makes the following contributions:

\begin{itemize}
    \item The problem of attribute missing graph is studied, and a novel method named AmGCL is proposed that integrates node attribute imputation and network embedding into a unified model.
    \item AmGCL applies Dirichlet energy minimization to feature precoding and reconstruction losses, which enhances the model's interpolation capabilities.
    \item Unlike previous methods based on generative frameworks, we introduce a GACLS that establishes a mutual information maximization \cite{shannon2001mathematical} objective function to supervise the generation of latent variables.
    \item Extensive experiments on real-world datasets demonstrate that our method outperforms other methods significantly in feature imputation and downstream node classification. Additionally, we also introduce a variant of AmGCL that achieves optimal classification performance by fine-tuning the model.
\end{itemize}

\section{Related Work}\label{Sec:II} 
Recent research has focused on addressing the problem of missing attributes in graph community detection using both traditional and deep learning-based methods. 

In the traditional methods, singular value threshold algorithms have been proposed to complete the matrix by Cai \emph{et al.} \cite{cai2010singular}, while Candes \emph{et al.} \cite{candes2012exact} have used convex optimization to accurately complete the matrix. Additionally, Hastie \emph{et al.} \cite{hastie2015matrix} have proposed a fast alternating least squares method to tackle this challenge.

Several deep learning-based methods have been proposed to address the issue of missing node attributes. Vincent \emph{et al.} \cite{vincent2010stacked} utilized denoising autoencoders to extract robust features, while Yoon \emph{et al.} \cite{yoon2018gain} applied GAN to data aggregation to generate missing data based on the true data distribution. Huang  \emph{et al.} \cite{huang2019graph} and Chen \emph{et al.} \cite{chen2019attributed} utilized attributed random walk techniques to generate node embeddings on bipartite graphs with node attributes. Spinelli \emph{et al.} \cite{spinelli2020missing} used a graph denoising autoencoder with each edge encoding the similarity between patterns to complete missing attributes. Taguchi \emph{et al.} \cite{taguchi2021graph} transformed missing attributes into Gaussian mixture distributions to make GCN applicable to incomplete network attributes. Xu \emph{et al.} \cite{xu2021adversarial} sought a common latent representation for multi-view data to infer missing data. Jin \emph{et al.} \cite{jin2022amer} used two subnetworks to process node attributes and graph structure and complemented data using structural information guided by the shared latent space assumption. Advanced methods like SAT \cite{chen2022learning} and SVGA \cite{yoo2022accurate} utilized separate subnetworks for node attributes and graph structure and used GAN and GRMF, respectively, to complete missing data with structural information guided by shared latent space assumptions.

These related works demonstrate the importance and complexity of completing missing attributes in graphs, and highlight the need for novel and effective methods to address this problem. AmGCL provides a promising approach that integrates both node attribute completion and network embedding into a unified model using a contrastive learning framework.

\section{Method Overview}\label{Sec:III} 

AmGCL is a graph neural network model designed to address the problem of attribute missing graph data. The model leverages self-supervised graph augmentation contrastive learning to improve performance. Figure 3 illustrates the overall framework, which includes five main parts: (a) preliminary, (b) feature precoding, (c) attribute graph augmentation, (d) GACLS, and (e) loss optimization. 
\begin{figure*}[htbp]
\centerline{\includegraphics[width=15cm]{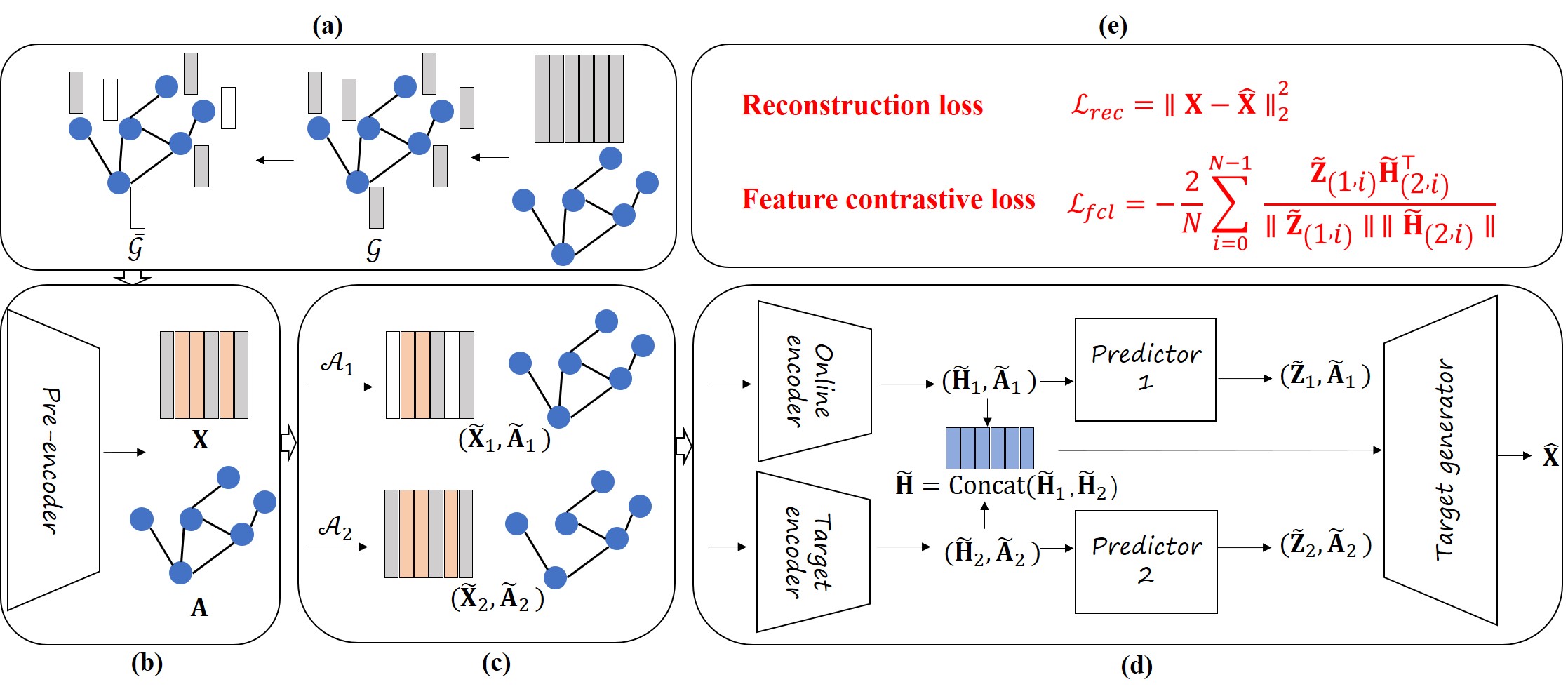}}
\caption{The AmGCL framework. Firstly, the  attribute missing graph $\bar{\mathcal{G}}$ is input into the precoder for feature encoding, resulting in a feature vector $(\mathbf{X},\mathbf{A})$. Then, using the augmented distribution method, the augmented features $(\widetilde{\mathbf{X}}_1,\widetilde{\mathbf{A}}_1)$ and $(\widetilde{\mathbf{X}}_2,\widetilde{\mathbf{A}}_2)$ are obtained and input into the GACLS module. In the GACLS module, the augmented features are encoded using the online encoder and target encoder, resulting in latent features $\widetilde{\mathbf{H}}_1$ and $\widetilde{\mathbf{H}}_2$, respectively. These features are then merged into $\widetilde{\mathbf{H}}$, which is decoded by the target generator to obtain the reconstructed node attribute features $\hat{\mathbf{X}}$.}
\label{fig_1}
\end{figure*}

In the feature precoding, the model employs the Dirichlet energy minimisation method to predict missing node attributes for subsequent graph representation learning. In the attribute graph augmentation, the model uses graph augmentation methods to enhance the robustness and generalisation of the model. The GACLS module uses self-supervised graph contrastive learning methods to learn the latent variable representations of nodes, and target generator to reconstruct the features. Finally, in the loss optimization, the model employs reconstruction loss and feature contrastive loss function to improve contrast sensitivity and model robustness during training.
\vspace{-2mm}
\subsection{Preliminary}
Observing from Figure \ref{fig_1} (a), we define the attribute graph $\mathcal{G}$, the set of nodes is denoted as $\mathcal{V(G)}=\{v_1,...,v_n\}$, and the set of edges as $\mathcal{E(G)}=\{e_1,...,e_n\}$. Edge $e_{ij}$ connects nodes $v_i$ and $v_j$, and $\mathcal{N}(i)=\{v_j:(v_i,v_j)\in \mathcal{E}\}$ represents the neighbor set of node $v_i$. Each node has a $d$-dimensional feature vector $\mathbf{X}\in \mathbb{R}^{N \times d}$ representing its attribute features. The attribute missing graph $\bar{\mathcal{G}}$ has two sets of nodes: $\mathcal{V}^o=\{v_1^o,v_2^o,...,v_{N_o}^o\}$ and $\mathcal{V}^m=\{v_1^m,v_2^m,...,v_{N_m}^m\}$, representing the sets of nodes with observed and missing attributes, respectively. The node set in the graph is denoted as $\mathcal{V}=\mathcal{V}^o\cup \mathcal{V}^m$, with $\mathcal{V}^o$ and $\mathcal{V}^m$ being disjoint sets, and $N=N_o+N_m$. The attribute feature vectors for the observed and missing node sets are denoted by $\mathbf{X}^o$ and $\mathbf{X}^m$, respectively. We aim to perform feature imputation and node classification on graphs with missing attributes in this paper. Our method involves using the available observed attribute values to interpolate the missing ones, while ensuring that the resulting embedding is suitable for the subsequent node classification task.

\subsection{Feature Precoding}
In attribute graphs, the node's attribute features are usually represented as high-dimensional vectors. Directly inputting missing attribute features into the network may cause network redundancy and introduce noise. Many feature imputation methods have been proposed, including imputation, matrix completion-based methods \cite{xu2015multi}, and graph-based methods \cite{chen2022learning}. In this paper, we introduce a feature precoding method based on Dirichlet energy minimization \cite{jeffreys1999methods}. The advantage is that it can effectively estimate missing features by utilizing the structural similarity and connectivity between nodes.

As shown in Figure \ref{fig_1} (b), before performing feature precoding, the features of each node need to be initialized. In this paper, the node's feature vector is initialized with its true feature, and for missing nodes, it is initialized with a zero vector. Next, for each node $v_i$, the contribution of its neighboring nodes $v_j \in \mathcal{N}(v_i^o)$ to its features is calculated, and these contributions are accumulated to obtain a new feature vector $\mathbf{X}_i$, which represents the feature of node $i$ after one round of feature precoding. This process can be calculated by:
\begin{equation}
\mathbf{X}_i^{(l+1)}=\mathbf{X}_i^{(l)}+\sum_{j \in \mathcal{N}\left(v_i^o\right)} w_{i j}\left(\mathbf{X}_j^{(l)}-\mathbf{X}_i^{(l)}\right)
\end{equation}
where $w_{ij}$ is a normalization factor that balances the degree difference between different nodes, and $l$ represents the number of iterations. Generally, multiple rounds of feature precoding can result in more accurate node features. Therefore, we iterate the above steps multiple times until convergence or until the predetermined number of iterations is reached. In this paper, the number of iterations is set to 40.
\vspace{-2mm}
\subsection{Attribute Graph Augmentation}
When an attribute graph contains missing values, node features may not fully reflect the true state of the nodes. Therefore, attribute graph augmentation can improve learning performance on attribute graphs by adding meaningful edges or removing irrelevant edges to improve the graph structure.

As shown in Figure \ref{fig_1} (c), by applying two random graph augmentation functions $\mathcal{A}_1$ and $\mathcal{A}_2$ to produce two alternative views of $\mathcal{G}$: 

\begin{equation}
\begin{aligned}
&\left({\widetilde{\mathbf{X}}}_1,{\widetilde{\mathbf{A}}}_1\right) = \mathcal{A}_1(\mathbf{X}_1, \mathbf{A}_1),\\
&\left({\widetilde{\mathbf{X}}}_2,{\widetilde{\mathbf{A}}}_2\right) = \mathcal{A}_2(\mathbf{X}_2, \mathbf{A}_2).
\end{aligned}
\end{equation}

Attribute graph augmentation can be considered in two ways: feature masking \cite{you2020graph} and edge dropout \cite{zhu2020deep}. Feature masking can reduce the interference of noise on the model by masking some node features. Edge dropout can reduce noise interference and redundant information by deleting edges, thereby improving the quality and stability of node features. Specifically, in the attribute graph missing problem, feature masking and edge dropout can remove noise edges that may interfere with learning, reducing the negative impact of attribute graph missing on node feature learning. These augmentations are graph-level and do not operate on each node independently, leveraging graph topology information.

\subsection{GACLS}
To learn the informative latent representation of the data and improve the model's generalization performance, we propose a GACLS as the training network (Figure \ref{fig_1} (d)), inspired by BYOL \cite{grill2020bootstrap}. GACLS uses two graph encoders: an online encoder $E_\theta$ and a target encoder $E_\phi$, with different parameters $\theta$ and $\phi$. The online encoder generates an online representation, $\widetilde{\mathbf{H}}1 := E_\theta(\widetilde{\mathbf{X}}_1,\widetilde{\mathbf{A}}_1)$, from the first augmented graph, while the target encoder generates a target representation, $\widetilde{\mathbf{H}}2 := E_\phi(\widetilde{\mathbf{X}}_2,\widetilde{\mathbf{A}}_2)$, from the second augmented graph. The output of the online network is passed through a predictor to obtain a prediction vector, and the output of the target network is used as the target vector for the predictor, updated via a slow-moving average to make the online and target network outputs approach each other in latent space. The online representation is then inputted into the predictor $p_\theta$, which outputs a prediction for the online representation, $\widetilde{\mathbf{Z}}_1 := p_\theta(\widetilde{\mathbf{H}}_1)$. It is worth noting that the predictor works at the node level and does not use graph information (i.e., it only operates on $\widetilde{\mathbf{H}}_1$ and not $\widetilde{\mathbf{A}}_1$). Similarly, the prediction of the target representation is $\widetilde{\mathbf{Z}}_2 := p_\theta(\widetilde{\mathbf{H}}_2)$.

By combining GACLS with the attribute graph feature missing problem, we train the online and target networks to learn latent vector representations for feature prefilling of missing node attributes. These vectors are then merged and inputted into a target generator to reconstruct the missing features.

\subsection{Loss Optimization}
AmGCL requires an effective loss function to guide model training. To improve model learning, AmGCL employs a comprehensive loss function that comprises two components, namely reconstruction loss and feature contrastive loss ( Figure \ref{fig_1} (e)). The reconstruction loss is guided by the principle of Dirichlet energy minimisation and is used to guide the model on known attribute features, aiming to enable the model to better predict the attribute features of missing nodes. On the other hand, the feature contrastive loss aims to maximise the lower bound on mutual information evidence and is used to guide the model in learning the similarity of feature representations to improve the model's performance in node classification tasks.

\subsubsection{Feature Contrastive Loss}

To bring the prediction closer to the true target while avoiding model collapse, we update the online parameter $\theta$ using the gradient of the cosine similarity relative to the online encoder, while the parameter $\phi$ of the target encoder is updated using an exponential moving average of the online parameter $\theta$. The feature contrastive loss function can be expressed as:
\begin{equation}\label{e4}
\mathcal{L}_{\text{fcl}}=-\frac{2}{N} \sum_{i=0}^{N-1} \frac{p\left(\widetilde{\mathbf{H}}_{(1, i)}\right) \widetilde{\mathbf{H}}_{(2, i)}^{\top}}{\left\|p\left(\widetilde{\mathbf{H}}_{(1, i)}\right)\right\|\left\|\widetilde{\mathbf{H}}_{(2, i)}\right\|},
\end{equation}
where $\widetilde{\mathbf{H}}_1^i$ represents the representation of input data $\mathbf{X}_i$ obtained through the online network, while $\widetilde{\mathbf{H}}_2^i$ represents the representation of input data $\mathbf{X}_i$ obtained through the target network. $p(\cdot)$ is the predictor, and $N$ is the batch size. These two representations are concatenated into an embedding representation $\widetilde{\mathbf{H}}$, which is defined as $\widetilde{\mathbf{H}} :=[\widetilde{\mathbf{H}}_1,\widetilde{\mathbf{H}}_2]$, where $[\cdot,\cdot]$ denotes the vector concatenation operation. By minimizing the loss functions of the predictor network and the target network, AmGCL can learn the latent representation of the data and apply it to downstream tasks. AmGCL's performance on the Steam dataset is poor due to the low network connectivity, which makes it difficult for the model to capture attributes through the network structure.

\subsubsection{Reconstruction Loss}
Feature reconstruction is a widely used training technique in self-supervised learning, where the original feature vector is first mapped to another vector and then reconstructed back through inverse mapping. The reconstruction loss, which employs Dirichlet energy minimisation, measures the difference between the reconstructed feature and the original feature. Its objective is to minimize the reconstruction error of the network, so that the learned features can retain the information of the original features and be utilized for downstream tasks.

To be specific, we feed the latent feature vector $\mathbf{H}$ into a target generator $D_\mu$ composed of a multi-layer perceptron (MLP). The target generator aims to map the feature vector to another vector $\hat{\mathbf{X}} := D_\mu(\mathbf{H})$ and then reconstruct it back through inverse mapping. Therefore, we employ the mean squared error (MSE) as the loss function between the online network and the reconstructor, which can be represented as:
\begin{equation}\label{e5}
\mathcal{L}_{\text{rec}}=\frac{1}{N} \sum_{i=0}^{N-1}\left\|\mathbf{X}_i-\widehat{\mathbf{X}}_i\right\|_2^2.
\end{equation}

The comprehensive loss function can be represented as:
\begin{equation}
\mathcal{L} = \mathcal{L}_{\text{rec}} + \lambda \mathcal{L}_{\text{fcl}},
\end{equation}
where $\lambda$ is the hyperparameter that controls the weight of the two loss components. Detailed theoretical proofs can be found in Appendix A.

\section{Experimentation and analysis}\label{Sec:IV} 
In this section, we address the following research questions:

Q1: How does AmGCL perform on feature imputation tasks? (Section 4.4)

Q2: What are the results of AmGCL on node classification tasks? (Section 4.5)

Q3: How does AmGCL generalize and perform under different rates of missing input data? (Section 4.6.1)

Q4: What is the role of each module in AmGCL, as evaluated through ablation studies? (Section 4.6.2)

Q5: What is the contribution of the two loss functions to the performance of AmGCL, as evaluated through ablation studies? (Section 4.6.3)

Q6: What is the computational complexity of AmGCL and how does it compare to other state-of-the-art methods? (Section 4.7)
\subsection{Datasets}
We have selected six publicly available graph datasets as the experimental objects in this study, each of which contains important information such as nodes, edges, features and subgraphs. The graph structure and node features of these datasets have been widely used in various tasks such as node classification and representation learning. We conducted a brief data statistics and analysis, and the specific situation is shown in Table \ref{tab_1}.

\begin{table}[htbp]\scriptsize
  \centering
  \caption{Data statistics.}
    \begin{tabular}{cccccc}
    \hline
    Dataset &{Nodes} & {Edges} & {Feature} & {Classes} & {Subgraphs} \\
    \hline
    Cora & 2,708 & 10,556 & 1,433 & 7 & 78\\
    CiteSeer & 3,327 & 9,104 & 3,703 & 6 & 438\\
    PubMed & 19,717 & 88,648 & 500 & 3 & 1\\
    CS & 18,333 & 163,788 & 6,805 & 15 & 1\\
    Computers & 13,752 & 491,722 & 767 & 10& 314 \\
    Photo & 7,650 & 238,162 & 745 & 8 & 136\\
    Seam & 9,944 & 266,981 & 352 & - & 6,741\\
    \hline
    \end{tabular}%
  \label{tab_1}%
\end{table}%
The Cora dataset \cite{yang2016revisiting} is a widely used benchmark for evaluating the performance of graph-based machine learning algorithms.

The CiteSeer dataset \cite{yang2016revisiting} is a digital library and search engine focused on computer and information science literature. The CiteSeer dataset is often used to evaluate graph-based machine learning algorithms.

The PubMed dataset \cite{yang2016revisiting} is a benchmark for evaluating machine learning algorithms for scientific literature analysis. 

The CS dataset \cite{shchur2018pitfalls} is based on the Microsoft Academic Graph co-authorship network, using data provided by the KDD Cup 2016 challenge. In this dataset, each node represents an author, and if two authors co-authored a paper, there is an edge connecting these two nodes. The node features represent the paper keywords for each author, while the labels indicate the author's most active research area.

The Computers and Photo datasets \cite{shchur2018pitfalls} are fragments of the Amazon co-purchasing graph, where nodes represent products and edges indicate frequent co-purchasing of two products. The node features are bag-of-words encoded product reviews, and the labels are given by the product categories.

The Steam dataset \cite{chen2022learning} contains 9944 games and 352 tags. The co-purchasing frequency between every two games was calculated based on user purchasing behavior, and a sparse item co-purchasing graph was obtained by binarizing the co-purchasing frequency using a threshold of 10.
\begin{table*}[htbp]\scriptsize
  \centering
  \caption{Feature imputation results on attribute missing graph with 60\% missing data.}
\renewcommand\arraystretch{1.5}
   \renewcommand\tabcolsep{2.1pt}
    \begin{tabular}{cccccccccccccccccc}
    \hline
    \multirow{1}[4]{*}{Metric} & \multirow{1}[4]{*}{Method} & \multirow{1}[4]{*}{Type}& \multicolumn{3}{c}{Cora} & \multicolumn{3}{c}{CiteSeer} & \multicolumn{3}{c}{Computers} & \multicolumn{3}{c}{Photo} & \multicolumn{3}{c}{Steam} \\
\cline{4-17}      &  & & @10 & @20 & @50 & @10 & @20 & @50 & @10 & @20 & @50 & @10 & @20 & @50 & @3 & @5 & @10 \\
    \hline
    \multirow{9}[2]{*}{Recall} & NeighAggre & Aggregation & 0.0906  & 0.1413  & 0.1961  & 0.0511  & 0.0908  & 0.1501  & 0.0321  & 0.0593  & 0.1306  & 0.0329  & 0.0616  & 0.1361  & 0.0603  & 0.0881  & 0.1446  \\
      & VAE & Generation & 0.0887  & 0.1228  & 0.2116  & 0.0382  & 0.0668  & 0.1296  & 0.0255  & 0.0502  & 0.1196  & 0.0276  & 0.0538  & 0.1279  & 0.0564  & 0.0820  & 0.1251  \\
      & GNN* & Representation& 0.1350  & 0.1812  & 0.2972  & 0.0620  & 0.1097  & 0.2058  & 0.0273  & 0.0533  & 0.1278  & 0.0295  & 0.0573  & 0.1324  & 0.2395  & 0.3431  & 0.4575  \\
      & GraphRNA & Generation & 0.1395  & 0.2043  & 0.3142  & 0.0777  & 0.1272  & 0.2271  & 0.0386  & 0.0690  & 0.1465  & 0.0390  & 0.0703  & 0.1508  & 0.2490  & 0.3208  & 0.4372  \\
      & ARWMF & Generation& 0.1291  & 0.1813  & 0.2960  & 0.0552  & 0.1015  & 0.1952  & 0.0280  & 0.0544  & 0.1289  & 0.0294  & 0.0568  & 0.1327  & 0.2104  & 0.3201  & 0.4512  \\
      & FP & Aggregation & 0.1571  & 0.2224  & 0.3338  & 0.0851  & 0.1368  & 0.2283  & 0.0425  & 0.0741  & 0.1544  & 0.0434  & 0.0773  & 0.1627  & 0.2095  & 0.2733  & 0.3664  \\
      & SAT & Generation& 0.1653  & 0.2345  & 0.3612  & 0.0811  & 0.1349  & 0.2431  & 0.0421  & 0.0746  & 0.1577  & 0.0427  & 0.0765  & 0.1635  & 0.2536  & 0.3620  & 0.4965  \\
      & SVGA & Generation& \underline{0.1718}  & \underline{0.2486}  & \underline{0.3814} & \underline{0.0943}  & \underline{0.1539 } & \underline{0.2782}  & \underline{0.0437}  & \underline{0.0769}  & \underline{0.1602}  & \underline{0.0446}  &\underline{0.0798}  & \textbf{0.1670} & \underline{0.2565}  & \underline{0.3602}  & \underline{0.4996}  \\
      & AmGCL & Contrast&\textbf{0.1811} & \textbf{0.2536} & \textbf{0.3852}  & \textbf{0.1031} & \textbf{0.1636} & \textbf{0.2819} & \textbf{0.0441} & \textbf{0.0769} & \textbf{0.1603} & \textbf{0.0449} & \textbf{0.0798} & \underline{0.1669}  & \textbf{0.2673} & \textbf{0.3621} & \textbf{0.5067} \\
    \hline
    \multirow{9}[2]{*}{NDCG} & NeighAggre& Aggregation & 0.1217  & 0.1548  & 0.1850  & 0.0823  & 0.1155  & 0.1560  & 0.0788  & 0.1156  & 0.1923  & 0.0813  & 0.1196  & 0.1998  & 0.0955  & 0.1204  & 0.1620  \\
      & VAE & Generation& 0.1224  & 0.1452  & 0.1924  & 0.0601  & 0.0839  & 0.1251  & 0.0632  & 0.0970  & 0.1721  & 0.0675  & 0.1031  & 0.1830  & 0.0902  & 0.1133  & 0.1437  \\
      & GNN* &Representation& 0.1736  & 0.2076  & 0.2702  & 0.1026  & 0.1423  & 0.2049  & 0.0671  & 0.1027  & 0.1824  & 0.0705  & 0.1082  & 0.1893  & 0.3366  & 0.4138  & 0.4912  \\
      & GraphRNA & Generation& 0.1934  & 0.2362  & 0.2938  & 0.1291  & 0.1703  & 0.2358  & 0.0931  & 0.1333  & 0.2155  & 0.0959  & 0.1377  & 0.2232  & 0.3437  & 0.4023  & 0.4755  \\
      & ARWMF & Generation& 0.1824  & 0.2182  & 0.2776  & 0.0859  & 0.1245  & 0.1858  & 0.0694  & 0.1053  & 0.1851  & 0.0727  & 0.1098  & 0.1915  & 0.3066  & 0.3877  & 0.4704  \\
      & FP & Aggregation& 0.2220  & 0.2668  & 0.3249  & 0.1409  & 0.1861  & 0.2445  & 0.1051  & 0.1473  & 0.2332  & 0.1061  & 0.1508  & 0.2411  & 0.2674  & 0.2377  & 0.2855  \\
      & SAT & Generation& 0.2250  & 0.2723  & 0.3394  & 0.1385  & 0.1834  & 0.2545  & 0.1030  & 0.1463  & 0.2346  & 0.1047  & 0.1498  & 0.2421  & 0.3585  & \textbf{0.4400} & \underline{0.5272}  \\
      & SVGA & Generation& \underline{0.2381}  & \underline{0.2894}  & \underline{0.3601}  & \underline{0.1579}  & \underline{0.2076}  &\underline{0.2892}  & \underline{0.1068}  & \underline{0.1509}  & \underline{0.2397}  & \underline{0.1084}  & \underline{0.1549}  & \textbf{0.2472} & \underline{0.3567}  & \underline{0.4391}  & \textbf{0.5299} \\
      & AmGCL& Contrast& \textbf{0.2507} & \textbf{0.2979} & \textbf{0.3644} & \textbf{0.1734} & \textbf{0.2239} & \textbf{0.3019} & \textbf{0.1082} & \textbf{0.1516} & \textbf{0.2404} & \textbf{0.1090} & \textbf{0.1549} & \underline{0.2469}  & \textbf{0.3595} & 0.4222  & 0.5011 \\
    \hline
    \end{tabular}%
  \label{tab_2}%
\end{table*}%

\subsection{Baselines}
We compared AmGCL with several existing baseline models in our study. NeighAgg \cite{csimcsek2008navigating} is a classic method that aggregates the features of adjacent nodes by mean pooling. For the case of missing neighbor attributes, we exclude them from the node aggregation. We only use one-hop neighbors as node neighbors. VAE \cite{kingma2013auto} is a generative model that learns the latent representation of instances. For test nodes with no attributes, we use neighbor aggregation methods (e.g., NeighAggre) to obtain the aggregated latent code of the test node in the latent space. Then, we use the decoder in VAE to reconstruct the missing node attributes. GCN \cite{kipf2016semi}, GraphSAGE \cite{hamilton2017inductive}, and GAT \cite{velivckovic2017graph} are popular graph neural networks that have been widely applied in various domains. For simplicity, we refer to the best-performing models in these three models as GNN*. GraphRNA \cite{huang2019graph} and ARWMF \cite{chen2019attributed} are recent representation learning methods that can be used for feature generation. FP \cite{rossi2022unreasonable} represents the feature precoding method, which uses Dirichlet energy minimization to fill in missing features. SAT \cite{chen2022learning} uses a shared latent space to train independent autoencoders that provide a shared latent space for features and graph structure, respectively. SVGA \cite{yoo2022accurate} is a state-of-the-art feature estimation model that establishes the feature space of features and graph structure by introducing Gaussian Markov random fields.

We divided AmGCL into two versions. AmGCL* is a variant that only uses $\mathcal{L}_{\text{fcl}}$ as the loss function and uses $\hat{\mathbf{X}}$ as the input for downstream classification. AmGCL's loss function is $\mathcal{L}_{\text{fcl}}+\mathcal{L}_{\text{rec}}$, and it uses $\widetilde{\mathbf{H}}$ as the input for downstream classification. 

\subsection{Experimental Setup}
AmGCL uses masking to represent missing feature attributes. Unless otherwise stated, we treat 40\% of the observable data as the training set and consider 60\% of the attribute missing nodes as target nodes (missing attribute nodes) and split them into validation and test sets in a 1:5 ratio, as in previous work \cite{yoo2022accurate}. We employ different experimental procedures for feature imputation and node classification.

For feature imputation, the latent layer dimension of each GNN and linear layer is set to 128. We run each experiment 10 times and report the average results. Each experiment runs for a maximum of 1000 epochs, and we use the Adam optimizer with a learning rate of 0.001 to train the model for the best results.

For node classification, we use the target nodes and their induced subgraphs (i.e., edges that only involve the target nodes) and divide the data set into training and testing sets in an 8:2 ratio, performing 5-fold cross-validation on the target nodes. We run a maximum of 1000 epochs and evaluate the quality of the generated features based on the accuracy of node classification. We use GCN as the classifier, with 2 layers, a dropout rate of 0.2, and a latent layer dimension of 256. We also use the Adam optimizer with a learning rate of 0.001 to train the model.

\subsection{Feature Imputation Results}
High-quality reconstructed attributes should have a high probability similar to that of real attributes in a given dimension. We conducted experiments on missing features and used Recall@k and NDCG@k as evaluation metrics for the discrete feature dataset. Specifically, we set k = \{10, 20, 30\} for Cora, CiteSeer, Computers, and Photo datasets, and k = \{3, 5, 10\} for the Steam dataset. The results of feature estimation are presented in Table \ref{tab_2}. By comparing with the baseline, we observed that our method outperformed the others in most cases.

Based on the evaluation results, NeighAggre and VAE performed the worst as they failed to achieve robust learning results at the attribute level, indicating that they were not effective in representing node features. GCN achieved good results, indicating that it can better capture the correlations between node features and effectively learn node representations. Random walk based attribute aggregation methods such as GraphRNA and ARWMF can capture the correlations in the attribute dimensions, but the statistical noise introduced by random walks may lead to moderate results. SAT and SVGA, as advanced deep generative models, focus on solving the problem of missing attribute graphs and have achieved some competitiveness in the results. AmGCL performed the best in most cases because it uses a more effective network structure and loss function. We conducted an analysis to understand why AmGCL did not perform as well on Steam dataset, and found that poor connectivity within the graph hindered the propagation of features along the graph structure.

\subsection{Node Classification Results}
In the node classification task on the attribute missing graph, we select the target node features that perform the best on the Recall@10 metric for node classification. From the table, it can be seen that (1) methods such as VAE, GCN, GraphSAGE, GAT, and Hers perform poorly because they are not specifically designed for missing attribute graphs. (2) Methods such as GraphRNA, AEWMF, SAT, and SVGA, which are specifically designed for attribute missing graphs, achieve certain results, but our method performs the best.

From Table \ref{tab_3}, it can be seen that AmGCL* outperforms AmGCL and achieves the state of the art. The loss function in AmGCL* only includes $\mathcal{L}_{\text{fcl}}$ without the feature reconstruction part, achieving the SOTA in node classification. This is because the node features themselves already contain enough information to distinguish different node categories. Therefore, the results of the node classification task may not be significantly affected by feature reconstruction, and even feature reconstruction may have a negative impact on classification. However, AmGCL* may not be able to utilize the diversity of learning tasks with node features. Therefore, it is still necessary to combine $\mathcal{L}_{\text{fcl}}$ with $\mathcal{L}_{\text{rec}}$ to obtain good results in both classification performance and feature reconstruction. The role of the two loss functions will be further reported in the ablation experiments.

\begin{table}[htbp]\scriptsize
  \centering
  \caption{Node classification results.}
  \renewcommand\arraystretch{1.5}
   \renewcommand\tabcolsep{2.1pt}
    \begin{tabular}{ccccccc}
    \hline
     Method  & Cora &  CiteSeer  & PubMed & CS &Computers  & Photo \\
    \hline
    NeighAggre & 0.6494  & 0.5413  & 0.6564  & 0.8031  & 0.8715  & 0.9010  \\
    VAE & 0.3011  & 0.2663  & 0.4007  & 0.2335  & 0.4023  & 0.3781  \\
    GCN & 0.4387  & 0.4079  & 0.4203  & 0.2180  & 0.3974  & 0.3656  \\
    GraphSage & 0.5779  & 0.4278  & 0.4200  & 0.2335  & 0.4019  & 0.3784  \\
    GAT & 0.4525  & 0.2688  & 0.4196  & 0.2334  & 0.4034  & 0.3789  \\
    GraphRNA & 0.8198  & 0.6394  & 0.8172  & 0.8851  & 0.8650  & 0.9207  \\
    ARWMF & 0.8025  & 0.2764  & 0.8089  & 0.8347  & 0.7400  & 0.6146  \\
    SAT & 0.8579  & 0.6767  & 0.7439  & 0.8402  & 0.8766  & 0.9260  \\
    SVGA & 0.8533  & 0.6879  & 0.6287  & 0.9037  & 0.8845  & 0.9272  \\
    AmGCL & \underline{0.8848}  & \underline{0.7703}  & \underline{0.8180}  & \underline{0.9040}  & \underline{0.8870}  & \underline{0.9281}  \\
    AmGCL* & \textbf{0.9179} & \textbf{0.7932} & \textbf{0.8525} & \textbf{0.9232} & \textbf{0.8871} & \textbf{0.9320} \\
    \hline
    \end{tabular}%
  \label{tab_3}%
\end{table}%

\subsection{Ablation Experiments}
We comprehensively evaluate the performance and generalization ability of each module of the model from three aspects: data ablation, GACLS model ablation, and loss function ablation.
\subsubsection{Data ablation}
To validate the effectiveness of in AmGCL across different data scenarios, we trained the model under missing rates ranging from 0.2 to 0.8 and evaluated the performance in terms of node classification. The results are shown in Table \ref{tab_4}. ``AmGCL w/ RF'' denotes using both observable and newly generated features as output, while ``AmGCL w/o FP'' denotes not using the initial feature precoding module. From Table \ref{tab_4}, it can be observed that (1) AmGCL achieved the best performance in most cases, and its ablations outperformed SVGA. (2) Using both the original observable features and newly generated missing features as the output can improve the accuracy of node classification. (3) Removing data feature propagation from the feature pre-coding module may have a negative impact on the results.

\begin{table}[htbp]\scriptsize
  \centering
  \caption{Data ablation results.}
    \renewcommand\arraystretch{1.5}
   \renewcommand\tabcolsep{2.1pt}
    \begin{tabular}{cccccccc}
    \hline
    Method & Missing & Cora &  CiteSeer  & PubMed & {CS} & {Computers} & {Photo} \\
    \hline
    \multirow{4}[2]{*}{SVGA} & 0.2  & 0.9243  & 0.7018  & 0.6167  & 0.9133  & 0.8754  & 0.9318  \\
      & 0.4  & 0.9177  & 0.7277  & 0.5790  & 0.9101  & 0.8813  & 0.9314  \\
      & 0.6  & 0.9006  & 0.6574  & 0.6287  & 0.9054  & 0.8845  & 0.9272  \\
      & 0.8  & 0.8763  & 0.6423  & 0.6060  & 0.9009  & 0.8810  & 0.9281  \\
    \hline
    \multirow{4}[2]{*}{AmGCL} & 0.2  & 0.9498  & 0.8822  & 0.8739  & 0.9543  & 0.9095  & 0.9497 \\
      & 0.4  & 0.9443  & 0.8603  & 0.8671  & 0.9443  & 0.9049  & 0.9393  \\
      & 0.6  & 0.9347  & 0.8223  & 0.8614  & 0.9435  & 0.9015  & 0.9349  \\
      & 0.8  & 0.9180  & 0.7472  & \textbf{0.8566} & 0.9218  & 0.8951  & 0.9275  \\
    \hline
    \multirow{4}[2]{*}{AmGCL w/o FP} & 0.2  & 0.9471  & 0.8747  & 0.8650  & 0.9775  & 0.8934  & 0.9409  \\
      & 0.4  & 0.9408  & 0.8523  & 0.8559  & 0.9719  & 0.8881  & 0.9347  \\
      & 0.6  & 0.9305  & 0.8343  & 0.8322  & 0.9517  & 0.8879  & 0.9348  \\
      & 0.8  & 0.9121  & 0.7240  & 0.8034  & 0.9258  & 0.8814  & 0.9233  \\
    \hline
    \multirow{4}[2]{*}{AmGCL w/ RF} & 0.2  & \textbf{0.9653} & \textbf{0.9162} & \textbf{0.8726} & \textbf{0.9781} & \textbf{0.9346} & \textbf{0.9597}\\
      & 0.4  & \textbf{0.9612} & \textbf{0.9053} & \textbf{0.8609} & \textbf{0.9726} & \textbf{0.9276} & \textbf{0.9552} \\
      & 0.6  & \textbf{0.9542} & \textbf{0.8548} & \textbf{0.8518} & \textbf{0.9603} & \textbf{0.9200} & \textbf{0.9448} \\
      & 0.8  & \textbf{0.9232} & \textbf{0.7707} & 0.8435  & \textbf{0.9421} & \textbf{0.9048} & \textbf{0.9349} \\
    \hline
    \end{tabular}%
  \label{tab_4}%
\end{table}%

Table \ref{tab_5} presents the performance of different methods on the node classification task when the node feature missing rate is 0.95. Raw denotes the original data, SVGA and FP are the two comparative methods, and AmGCL is the proposed method in this paper. From the table, it can be seen that the accuracy of AmGCL is higher than SVGA and the original data on all datasets. The FP method performs well on some datasets, indicating that AmGCL has good generalization performance, i.e., it can maintain high accuracy even in extremely missing data scenarios. Meanwhile, the performance of SVGA on some datasets is worse than the original data, indicating that SVGA has poor generalization ability.
\begin{table}[htbp]\scriptsize
  \centering
  \caption{Data ablation experiments at a missing rate of 0.95.}
      \renewcommand\arraystretch{1.5}
   \renewcommand\tabcolsep{2.1pt}
    \begin{tabular}{ccccccc}
    \hline
     Method  & Cora &  CiteSeer  & PubMed & CS   & Computers& Photo \\
    \hline
    Raw & 0.6440  & 0.4481  & 0.6746  & 0.7354  & 0.8468  & 0.8841  \\
    SVGA & 0.4760  & 0.2185  & 0.6226  & 0.8864  & 0.8426  & 0.9089  \\
    FP & 0.8667  & 0.6360  & 0.8198  & 0.9160  & 0.8777  & 0.9119  \\
    AmGCL & \textbf{0.9054} & \textbf{0.6651} & \textbf{0.8354} & \textbf{0.9217} & \textbf{0.8790} & \textbf{0.9146} \\
    \hline
    \end{tabular}%
  \label{tab_5}%
\end{table}%

\subsubsection{GACLS Module Ablation}
Results of GACLS model ablation experiments are presented in Table \ref{tab_6}. ``w/o generator'' indicates removing the target generator structure and directly using $\widetilde{\mathbf{H}}$ as the output; ``w/o concat'' means using only $\widetilde{\mathbf{H}}_1$ as the output; ``w/o concat + generator'' means removing the generator structure and using both $\widetilde{\mathbf{H}}_1$ as the output;  ``w/o FM'' means not using feature masking for graph augmentation; ``w/o ER'' means not using edge removal for graph augmentation. The results clearly demonstrate that AmGCL outperforms the others in most cases.

\begin{table}[htbp]\scriptsize
  \centering
  \caption{Results of GACLS model ablation experiments.}
        \renewcommand\arraystretch{1.5}
   \renewcommand\tabcolsep{2.1pt}
    \begin{tabular}{ccccccc}
    \hline
    Type  & Cora &  CiteSeer  & PubMed & CS   & Computers   & Photo \\
    \hline
    w/o generator & 0.9070  & 0.7818  & 0.8488  & 0.9176  & 0.8991  & 0.9339  \\
    w/o concat & 0.9243  & 0.8170  & 0.8490  & 0.9417  & 0.8987  & 0.9348  \\
    w/o concat +generator & 0.8977  & 0.7731  & 0.8309  & 0.9145  & 0.8845  & 0.9285  \\
    w/o FM & 0.9265  & 0.8209  & 0.8526  & 0.9413  & 0.9031  & 0.9388  \\
    w/o ER & 0.9158  & 0.8197  & 0.8483  & 0.9411  & 0.8978  & 0.9348  \\
    AmGCL & \textbf{0.9347} & \textbf{0.8223} & \textbf{0.8614} & \textbf{0.9435} & \textbf{0.9088} & \textbf{0.9349} \\
    \hline
    \end{tabular}%
  \label{tab_6}%
\end{table}%

\subsubsection{Loss Function Ablation}
Tables \ref{tab_7} and \ref{tab_8} present the results of the loss ablation experiments on feature reconstruction. As observed from the tables, the feature reconstruction performance slightly decreases when the $L_{\text{fcl}}$ loss is removed, indicating the positive impact of $L_{\text{fcl}}$ loss on improving feature reconstruction. However, a significant improvement in the node classification metrics was observed when the $L_{\text{rec}}$ loss was further removed, suggesting that the $L_{\text{rec}}$ loss interferes with the node classification task. The feature interpolation is important and must not be overlooked. Therefore, the combination of $L_{\text{fcl}}$ and $L_{\text{rec}}$ can strike a balance between the node classification and feature reconstruction tasks, leading to better performance.
\begin{table}[b]\tiny
  \centering
  \caption{Reconstruction loss ablation results.}
          \renewcommand\arraystretch{1.5}
   \renewcommand\tabcolsep{2pt}
    \begin{tabular}{ccccccccccc}
    \hline
    \multirow{1}[4]{*}{Metric} & \multirow{1}[4]{*}{Method}  & \multicolumn{3}{c}{CiteSeer} & \multicolumn{3}{c}{Computers} & \multicolumn{3}{c}{Photo} \\
\cline{3-11}      &    &@10 & @20 & @50 & @10 & @20 & @50 & @10 &@20 & @50\\
    \hline
    \multirow{2}[1]{*}{Recall} &  w/o $\mathcal{L}_{\text{fcl}}$  & 0.0974  & 0.1585  & 0.2744  & 0.0422  & 0.0742  & 0.1560  & 0.0436  & 0.0776  & \textbf{0.1651 } \\
      & AmGCL & \textbf{0.1030} & \textbf{0.1627} & \textbf{0.2809 } & \textbf{0.0434} & \textbf{0.0762} & \textbf{0.1587} & \textbf{0.0442} & \textbf{0.0784 } & 0.1648  \\
    \multirow{2}[1]{*}{NDCG} &  w/o $\mathcal{L}_{\text{fcl}}$  & 0.1674  & 0.2166  & 0.2928  & 0.1030  & 0.1457  & 0.2331  & 0.1063  & 0.1514  & \textbf{0.2441} \\
      & AmGCL  & \textbf{0.1729} & \textbf{0.2229} & \textbf{0.3010 } & \textbf{0.1058 } & \textbf{0.1494 } & \textbf{0.2375} & \textbf{0.1072} & \textbf{0.1525} & 0.2438 \\
    \hline
    \end{tabular}%
  \label{tab_7}%
\end{table}%

\begin{table}[htbp]\scriptsize
  \centering
  \caption{Feature contrastive loss ablation results.}
            \renewcommand\arraystretch{1.5}
   \renewcommand\tabcolsep{2.1pt}
    \begin{tabular}{ccccccc}
    \hline
     Method  & Cora &  CiteSeer  & PubMed & CS  & Computers  & Photo \\
    \hline
    AmGCL & 0.9095  & 0.7971  & 0.8563  & 0.9017  & 0.8877  & 0.9304  \\
    w/o $\mathcal{L}_{\text{rec}}$ & \textbf{0.9542} & \textbf{0.8548} & \textbf{0.8518} & \textbf{0.9603} & \textbf{0.9200} & \textbf{0.9448} \\
    \hline
    \end{tabular}%
  \label{tab_8}%
\end{table}%

\subsection{Method Performance Validation}
To further evaluate the performance of the model, we compared the training time and computational resources required by the model, as shown in Table \ref{tab_9} below. The complexity analysis of the AmGCL is shown in Appendix B.

From the Table \ref{tab_9}, we can see that the number of training parameters required by the AmGCL method is much smaller than the other two methods on different datasets. For example, on the PubMed dataset, the AmGCL method has only 243,203 parameters, which is much less than the 4,195,309 parameters of the SAT method and the 5,301,519 parameters of the SVGA method. In addition, we can also see that the number of parameters required by the AmGCL method on the Cora, CiteSeer, and Photo datasets is also much less than the other two generative methods. This means that the AmGCL method can be trained and deployed faster while requiring less computational resources.

The Table \ref{tab_9} provides information on the training time required by the three methods on various datasets. It is evident that the AmGCL method outperforms the other two methods in terms of training time. Specifically, on the CiteSeer, Computers, and Photo datasets, the AmGCL method takes significantly less time to train than the other methods. 

More detailed experimental analysis can be found in Appendices C, D and E.
\begin{table}[htbp]\scriptsize
  \centering
  \caption{Model parameters and training time statistics.}
              \renewcommand\arraystretch{1.5}
   \renewcommand\tabcolsep{2.1pt}
    \begin{tabular}{cccccccc}
    \hline
    Type &  Method  & Cora &  CiteSeer  & PubMed & CS   & Computers   & Photo \\
    \hline
    \multirow{3}[2]{*}{Params} & SAT & 1,167,642 & 2,201,712 & 4,195,309 & 3,109,376 & 3,109,376 & 1,880,154 \\
      & SVGA & 1,136,060 & 1,877,245 & 5,301,519 & 6,560,215 & 3,826,728 & 2,240,166 \\
      & AmGCL & \textbf{602,499} & \textbf{1,473,923} & \textbf{243,203} & \textbf{2,667,395} & \textbf{347,523} & \textbf{338,563} \\
    \hline
    \multirow{3}[2]{*}{Times} & SAT & 3.5661 & 4.3805& 6.3210 & 8.5790 & 8.5791 & 5.7879 \\
      & SVGA & 3.3365 & 3.4780 & 4.3585 & 10.7354 & 10.6048 & 7.2488 \\
      & AmGCL & \textbf{2.9250} & \textbf{3.2108} &  \textbf{4.1859} &  \textbf{7.6305} & \textbf{7.8639} & \textbf{4.5042} \\
    \hline
    \end{tabular}%
  \label{tab_9}%
\end{table}%

\section{Conclusion}\label{Sec:V} 

This paper proposes an AmGCL method to address the node attribute missing problem in attribute missing graph data. AmGCL deals with the problem of missing node attributes by combining Dirichlet energy minimisation and contrastive learning methods, which has not been applied by other existing methods. AmGCL uses feature precoding and a GACLS to learn potential representations of missing nodes, which can improve the performance of attribute graph analysis tasks by filling in missing node attributes while preserving the overall graph structure. Experimental results demonstrate that on multiple real-world datasets, AmGCL outperforms state-of-the-art methods in terms of feature imputation and node classification. Future work will explore how well feature interpolation can be combined with downstream tasks to fully exploit attribute missing graph learning.

\bibliographystyle{ACM-Reference-Format}
\bibliography{sample-base}

\clearpage
\appendix

\section{Research Methods}

\subsection{Mutual Information Proof}

GACLS is inspired by BYOL, which can be treated as a kind of non-negtive-sampling contrastive learning. And contrastive learning is essentially maximizing the mutual information of positive sample pair.

For a specific subgraph $c$, take 1 positive sample graph or augment and $ N - 1 $ negative samples from N different distributions respectively, then the probability that $x_i$ is a positive sample is:

\begin{equation}
\begin{aligned} p\left(d=i \mid X, c\right) &=\frac{p\left(x_{i} \mid c\right) \prod_{l \neq i} p\left(x_{l}\right)}{\sum_{j=1}^{N} p\left(x_{j} \mid c\right) \prod_{l \neq j} p\left(x_{l}\right)} \\ &=\frac{\frac{p\left(x_{i} \mid c\right)}{p\left(x_{i}\right)}}{\sum_{j=1}^{N} \frac{p\left(x_{j} \mid c\right)}{p\left(x_{j}\right)}} \end{aligned}
\end{equation}

It can be seen from the following formula that the goal of contrastive learning optimization is to maximize the mutual information lower bound.

\begin{equation}
\begin{aligned} \mathcal{L}_{\mathrm{N}}^{\mathrm{opt}} &=-\underset{X}{\mathbb{E}} \log \left[\frac{\frac{p\left(x_{i} \mid c\right)}{p\left(x_{i}\right)}}{\frac{p\left(x_{i} \mid c\right)}{p\left(x_{i}\right)}+\sum_{x_{j} \in X_{\mathrm{ncg}}} \frac{p\left(x_{j} \mid c\right)}{p\left(x_{j}\right)}}\right] \\ &=\underset{X}{\mathbb{E}} \log \left[1+\frac{p\left(x_{i}\right)}{p\left(x_{i} \mid c\right)} \sum_{x_{j} \in X_{\text {neg }}} \frac{p\left(x_{j} \mid c\right)}{p\left(x_{j}\right)}\right] \\ & \approx \underset{X}{\mathbb{E}} \log \left[1+\frac{p\left(x_{i}\right)}{p\left(x_{i} \mid c\right)}(N-1) \underset{x_{j}}{\mathbb{E}} \frac{p\left(x_{j} \mid c\right)}{p\left(x_{j}\right)}\right] \\ &=\underset{X}{\mathbb{E}} \log \left[1+\frac{p\left(x_{i}\right)}{p\left(x_{i} \mid c\right)}(N-1)\right] \\ & \geq \underset{X}{\mathbb{E}} \log \left[\frac{p\left(x_{i}\right)}{p\left(x_{i} \mid c\right)} N\right] \\ &=-I\left(x_{i}, c\right)+\log (N) \end{aligned}
\end{equation}

\subsection{Dirichlet Energy Proofs}

The feature precoding method is based on Dirichlet energy minimization. The recursive relation can be deduced as follows. 

\begin{equation}
\begin{aligned}
\min_X \mathcal{L} =& \sum_{(i,j)\in \mathcal{E}} w_{ij}(x_i-x_j)^2 = \text{tr}(X^TLX) \\
& \text{s.t. } X_k=Z_k \\
\implies& \mathcal{L'} = LX = (I-\widetilde A)X = 0, \quad X_k=Z_k \\
\implies& X = \widetilde AX , \quad X_k=Z_k
\end{aligned}
\end{equation}

\section{Method Complexity Analysis}
Analyzing the time and space complexity of AmGCL as shown in Table \ref{tab_a3}, where the number of attributes is $F$, the number of edges $\vert \mathcal{E} \vert$, the number of nodes $\vert \mathcal{V} \vert$, the number of categories $c$, and the dimension of the latent variable $d$, it can be seen from the table that the time complexity of AmGCL is $O(F \vert \mathcal{E} \vert + (Fd + d^2)\vert \mathcal{V} \vert)$ and the space complexity is $O(\vert \mathcal{E} \vert + (d + F) \vert \mathcal{V} \vert + d^2 + Fd)$.
\begin{table}[htbp]
  \centering
  \caption{Complexity analysis on AmGCL.}
    \begin{tabular}{lll}
    \hline
    Module & Time & Space \\
    \hline
    Precoder * 1 & $O(F \vert \mathcal{E} \vert)$ & $O(\vert \mathcal{E} \vert+F \vert \mathcal{V} \vert)$ \\
    Augment * 2& $O(\vert \mathcal{E} \vert+\vert\mathcal{V}\vert)$ & $O(\vert \mathcal{E} \vert+F)$ \\
    Encoder * 2& $O((F+d)(\vert\mathcal{E}\vert+d\vert\mathcal{V}\vert))$ & $O(\vert \mathcal{E} \vert+d^2+d\vert \mathcal{V} \vert)$ \\
    Predictor * 2& $O(d^2\vert\mathcal{V}\vert)$ & $O(d^2)$ \\
    Generator * 1& $O(Fd\vert\mathcal{V}\vert)$ & $O(Fd+F\vert\mathcal{V}\vert)$\\
    \hline
    \end{tabular}%
  \label{tab_a3}%
\end{table}%

\section{Experiments on Rational Motivation of AmGCL}

\begin{table*}[htbp]\scriptsize
  \centering
  \caption{Experimental results on the proof of model motivation.}
    \renewcommand\arraystretch{1.5}
   \renewcommand\tabcolsep{2.1pt}
    \begin{tabular}{cccccccccccccc}
    \hline
    \multirow{1}[4]{*}{Metric} & \multirow{1}[4]{*}{Method} & \multicolumn{3}{c}{Cora} & \multicolumn{3}{c}{CiteSeer} & \multicolumn{3}{c}{Computers} & \multicolumn{3}{c}{Photo} \\
\cline{3-14}      &   & @10 & @20 & @50 & @10 & @20 & @50 & @10 & @20 & @50 & @10 & @20 & @50 \\
    \hline
    \multirow{7}[2]{*}{Recall} & BYOL & 0.1456  & 0.2118  & 0.3430  & 0.0794  & 0.1324  & 0.2472  & 0.0407  & 0.0727  & 0.1537  & 0.0421  & 0.0751  & 0.1601  \\
      & BYOL+One hot  & 0.1402  & 0.2073  & 0.3321  & 0.0705  & 0.1190  & 0.2205  & 0.0410  & 0.0730  & 0.1543  & 0.0410  & 0.0739  & 0.1578  \\
      & BYOL+Dirichlet & 0.1341  & 0.1909  & 0.2981  & 0.0597  & 0.0914  & 0.1563  & 0.0314  & 0.0545  & 0.1164  & 0.0313  & 0.0565  & 0.1202  \\
      & Precoder & 0.1571  & 0.2224  & 0.3338  & 0.0851  & 0.1368  & 0.2283  & 0.0425  & 0.0741  & 0.1544  & 0.0434  & 0.0773  & 0.1627  \\
      & SAT & 0.1653  & 0.2345  & 0.3612  & 0.0811  & 0.1349  & 0.2431  & 0.0421  & 0.0746  & 0.1577  & 0.0427  & 0.0765  & 0.1635  \\
      & SVGA & 0.1718  & 0.2486  & \textbf{0.3814 } & 0.0943  & 0.1539  & 0.2782  & 0.0437  & 0.0769  & 0.1602  & 0.0446  & 0.0798  & \textbf{0.1670 } \\
      & AmGCL & \textbf{0.1811} & \textbf{0.2507} & 0.3752  & \textbf{0.1031} & \textbf{0.1636} & \textbf{0.2819} & \textbf{0.0441} & \textbf{0.0769} & \textbf{0.1603} & \textbf{0.0449} & \textbf{0.0798} & 0.1669 \\
    \hline
    \multirow{7}[2]{*}{nDCG} & BYOL & 0.2040  & 0.2481  & 0.3172  & 0.1277  & 0.1719  & 0.2472  & 0.0996  & 0.1420  & 0.2282  & 0.1023  & 0.1461  & 0.2360  \\
      & BYOL+One hot  & 0.1942  & 0.2390  & 0.3048  & 0.1190  & 0.1595  & 0.2260  & 0.1005  & 0.1431  & 0.2299  & 0.0999  & 0.1436  & 0.2325  \\
      & BYOL+Dirichlet & 0.1954  & 0.2335  & 0.2905  & 0.1038  & 0.1305  & 0.1732  & 0.0784  & 0.1098  & 0.1764  & 0.0781  & 0.1119  & 0.1801  \\
      & Precoder & 0.2220  & 0.2668  & 0.3249  & 0.1409  & 0.1861  & 0.2445  & 0.1051  & 0.1473  & 0.2332  & 0.1061  & 0.1508  & 0.2411  \\
      & SAT & 0.2250  & 0.2723  & 0.3394  & 0.1385  & 0.1834  & 0.2545  & 0.1030  & 0.1463  & 0.2346  & 0.1047  & 0.1498  & 0.2421  \\
      & SVGA & 0.2381  & 0.2894  & 0.3601  & 0.1579  & 0.2076  & 0.2892  & 0.1068  & 0.1509  & 0.2397  & 0.1084  & 0.1549  & \textbf{0.2472} \\
      & AmGCL & \textbf{0.2507} & \textbf{0.2977} & \textbf{0.3641} & \textbf{0.1734} & \textbf{0.2239} & \textbf{0.3019} & \textbf{0.1082} & \textbf{0.1516} & \textbf{0.2404} & \textbf{0.1090} & \textbf{0.1549} & 0.2469 \\
    \hline
    \end{tabular}%
  \label{tab_a1}%
\end{table*}%

This section analyzes the design process of AmGCL through experiments to better illustrate our contribution. The pseudo code of AmGCL is shown in Algorithm \ref{a1}.
\begin{algorithm}
\caption{Training process of AmGCL.}\label{a1}
\LinesNumbered
        \KwIn{Attribute-missing graph $\bar{\mathcal{G}}$, adjacency matrix $\mathbf{A}$, and observable node features $\mathbf{X}^o$, Iteration number $I$.}
        \KwOut{Reconstructed features $\hat{\mathbf{X}}$, embedded features $\widetilde{\mathbf{H}}$.}
        Initialize the parameters of GACLS;\\
        Perform feature precoder on $\mathbf{X}^o$ to obtain $\mathbf{X}$;\\
        \For{ $i=1$ to $I$ }{
        Enhance $\mathbf{A}$ and $\mathbf{X}$ using graph enhancement algorithms $\mathcal{A}_1$ and $\mathcal{A}_2$ to obtain (${\widetilde{\mathbf{X}}}_1{,\ \widetilde{\mathbf{A}}}_1$) and (${\widetilde{\mathbf{X}}}_2{,\ \widetilde{\mathbf{A}}}_2$); \\
        Use the Online encoder and Target encoder of GACLS to calculate (${\widetilde{\mathbf{H}}}_1{,\ \widetilde{\mathbf{A}}}_1$) and (${\widetilde{\mathbf{H}}}_2{,\ \widetilde{\mathbf{A}}}_2$);\\
	Input (${\widetilde{\mathbf{H}}}_1{,\ \widetilde{\mathbf{A}}}_1$) and (${\widetilde{\mathbf{H}}}_2{,\ \widetilde{\mathbf{A}}}_2$) into the predictor to obtain (${\widetilde{\mathbf{Z}}}_1{,\ \widetilde{\mathbf{A}}}_1$) and (${\widetilde{\mathbf{Z}}}_2{,\ \widetilde{\mathbf{A}}}_2$);\\
	Combine ${\widetilde{\mathbf{H}}}_1,\ {\widetilde{\mathbf{H}}}_2$ to obtain $\widetilde{\mathbf{H}}$;\\
        Input $\widetilde{\mathbf{H}}$ into the generator to obtain $\hat{\mathbf{X}}$;\\
	Use Equations \ref{e4} and \ref{e5} for contrastive learning and reconstruction learning;\\
	Back propagation and update parameters;\\    

        }
Return $\hat{\mathbf{X}}$ and $\widetilde{\mathbf{H}}$;\\
\end{algorithm}

\textbf{Overall design motivation}: Firstly, we notice that the state-of-the-art methods, SAT and SVGA, both impose restrictions on latent variables during reconstruction, with SAT using GAN and SVGA using GMRF. Although this is effective in some cases, it can lead to overfitting in other cases due to overly strict limitations. In contrast, AmGCL uses self-supervised contrastive learning for feature interpolation, without explicitly modeling or restricting the latent variables, but rather through mutual information constraints. This makes AmGCL more flexible and robust, able to adapt to various data distributions and missing patterns.

\textbf{Model design}:
Pre-encoder: For attribute missing graphs, we focus on the fact that the original distribution of features is disrupted, so the setting of the initial distribution is crucial. To build the initial distribution of features, AmGCL adopts a pre-encoder based on minimizing the Dirichlet energy. This pre-encoder uses graph structural information to propagate features, obtaining the initial feature representation for each node. Compared to traditional methods such as random padding or taking the mean, the pre-encoder can make more full use of graph structural information, rather than just attribute features. It can also adopt different orders of neighbors by controlling the number of propagation layers. Since the pre-encoder has no training parameters, it can be easily embedded into the entire architecture of AmGCL.
\begin{figure}[t]
\centerline{\includegraphics[width=8cm]{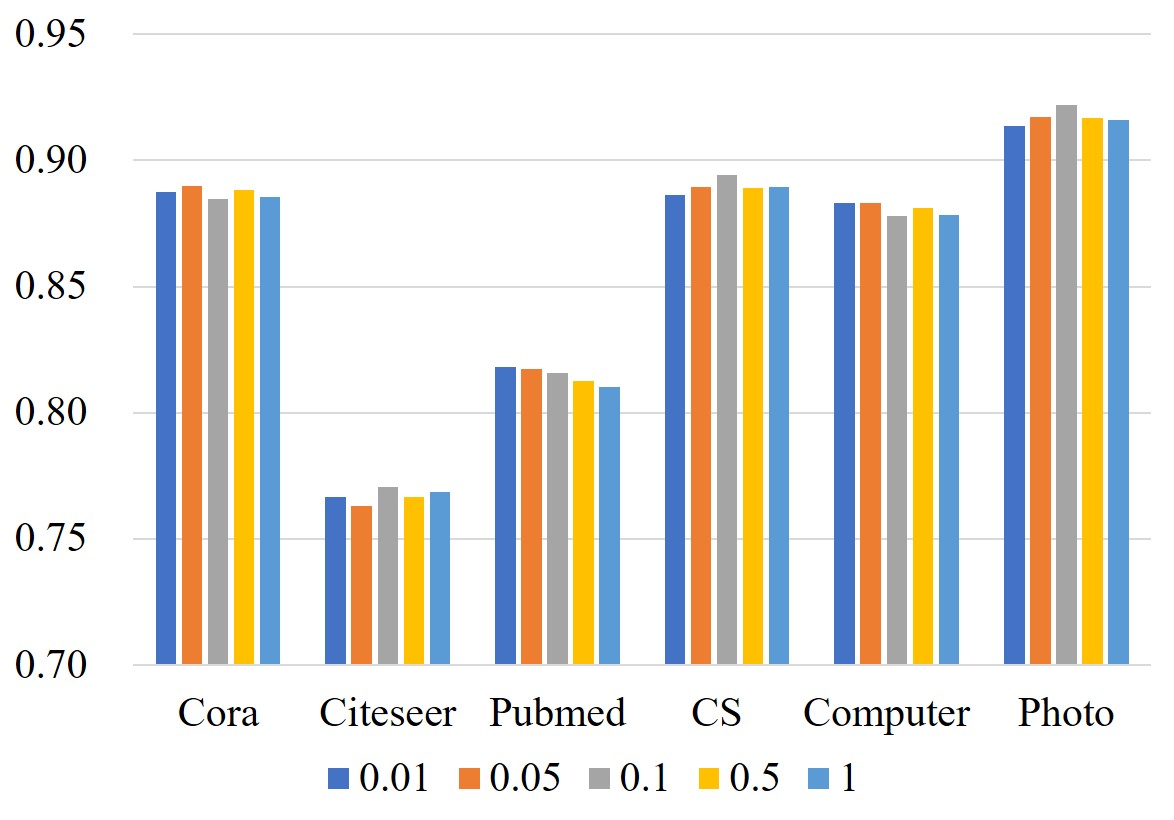}}
\caption{Equilibrium parameter $\lambda$ results.}
\label{fig_a3}
\end{figure}
GACLS: This architecture is the core of AmGCL, drawing on the idea of BYOL and using self-supervised contrastive learning for feature interpolation. Specifically, GACLS first encodes node features using an online and target encoder architecture, obtaining the latent feature representation. Then, by feeding the latent feature representation into the generator and removing the irrelevant components of mutual information based on energy minimization, the reconstructed feature representation is obtained.

We conducted experiments to fully demonstrate the effectiveness of the modules. From Table \ref{tab_a1}, ``BYOL'' represents using the BYOL framework, and initializing features as learnable parameters following a Gaussian distribution, training with both contrastive and reconstruction losses. ``BYOL+One hot'' feature represents initializing features as a one-hot encoding matrix on the basis of ``BYOL''. ``BYOL+Dirichlet'' represents incorporating Dirichlet energy minimization into the training loss on the basis of ``BYOL'', rather than representing it in the form of a pre-encoder. ``Precoder'' represents using only the pre-encoder for feature propagation as the final output. From the results, it can be seen that the performance of ``BYOL'' is significantly lower than that of AmGCL, because it requires more training parameters and has a larger search space. Both ``BYOL+One hot'' and ``BYOL+Dirichlet'' are inferior to using ``BYOL'' alone, which reversely proves the role of the pre-encoder. Meanwhile, ``Precoder'' has certain competitiveness, which also indicates that Dirichlet energy minimization is effective. The fact that AmGCL based on contrastive learning outperforms SAT and SVGA based on generative learning also demonstrates the effectiveness of self-supervised contrastive learning for attribute missing graph learning.

\begin{figure}
\subfigure[SAT]{
\includegraphics[width=0.2\textwidth]{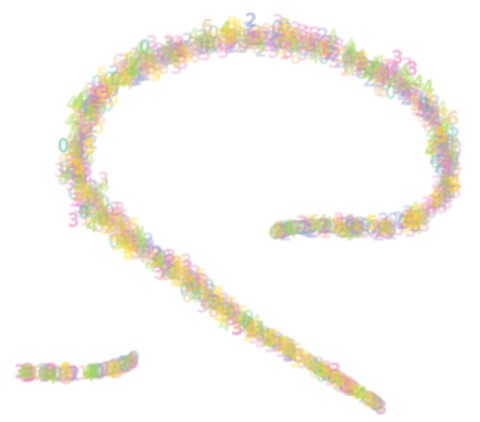}
}
\subfigure[SVGA]{
\includegraphics[width=0.2\textwidth]{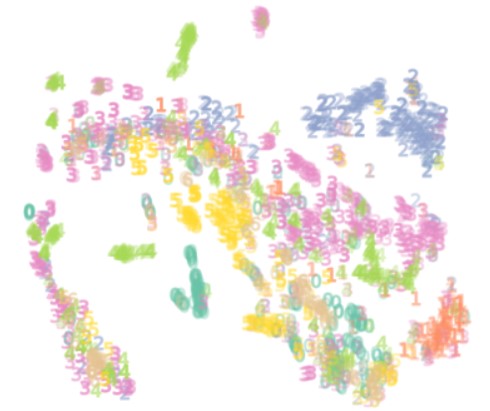}
}
\subfigure[AmGCL]{
\includegraphics[width=0.2\textwidth]{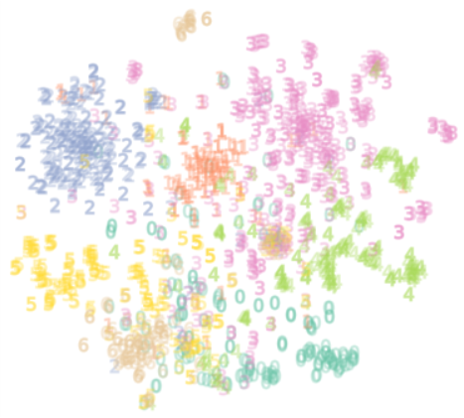}
}
\centering
\caption{Visual comparison of interpolation features.}
\label{fig_a1}
\end{figure}

\section{Hyperparameter Settings}
We have analysed the equilibrium parameters of the loss function and the results are shown in Figure \ref{fig_a3}.

\section{Interpolation Feature Visualization}

Since SAT and SVGA were the most advanced methods in previous experiments on attribute-missing graph learning, they were chosen as the most representative methods for comparison with AmGCL. We visualized the distribution of learned node embeddings for all three methods on all datasets using t-SNE \cite{van2008visualizing}. As shown in Figure \ref{fig_a1}, AmGCL exhibits a clearer structure and better reveals the intrinsic data structure compared to the baseline.

\end{document}